\documentclass[preprint,12pt]{elsarticle}

\usepackage{amssymb}
\usepackage{xcolor}
\journal{Journal of Manufacturing Systems}

\begin{document}

\begin{frontmatter}



\title{An interpretable deep learning method for bearing fault diagnosis}


\author[ISU-ME,ISU-ECE]{Hao Lu}
\author[UCONN]{Austin M. Bray}
\author[UCONN]{Chao Hu}
\author[Percev,Grace]{Andrew T. Zimmerman}
\author[UCONN]{Hongyi Xu}

\affiliation[ISU-ME]{organization={Department of Mechanical Engineering, Iowa State University},%Department and Organization
            %addressline={Address Two}, 
            city={Ames},
            postcode={50011}, 
            state={IA},
            country={USA}}

\affiliation[ISU-ECE]{organization={Department of Electrical and Computer Engineering, Iowa State University},%Department and Organization
            %addressline={Address Two}, 
            city={Ames},
            postcode={50011}, 
            state={IA},
            country={USA}}
            
\affiliation[UCONN]{organization={Department of Mechanical Engineering, University of Connecticut},%Department and Organization
            %addressline={}, 
            city={Storrs},
            state={CT},
            postcode={06269}, 
            country={USA}}

\affiliation[Percev]{organization={Perc$\bar{e}v$, LLC},%Department and Organization
            %addressline={Address Two}, 
            city={Davenport},
            postcode={52807}, 
            state={IA},
            country={USA}}

\affiliation[Grace]{organization={Grace Technologies},%Department and Organization
            %addressline={Address Two}, 
            city={Davenport},
            postcode={52807}, 
            state={IA},
            country={USA}}

\begin{abstract}

Deep learning (DL) has gained popularity in recent years as an effective tool for classifying the current health and predicting the future of industrial equipment. However, most DL models have black-box components with an underlying structure that is too complex to be interpreted and explained to human users. This presents significant challenges when deploying these models for safety-critical maintenance tasks, where non-technical personnel often need to have complete trust in the recommendations these models give. To address these challenges, we utilize a convolutional neural network (CNN) with Gradient-weighted Class Activation Mapping (Grad-CAM) activation map visualizations to form an interpretable DL method for classifying bearing faults. After the model training process, we apply Grad-CAM to identify a training sample's feature importance and to form a library of diagnosis knowledge (or health library) containing training samples with annotated feature maps. During the model evaluation process, the proposed approach retrieves prediction basis samples from the health library according to the similarity of the feature importance. The proposed method can be easily applied to any CNN model without modifying the model architecture, and our experimental results show that this method can select prediction basis samples that are intuitively and physically meaningful, improving the model's trustworthiness for human users. 

% In response, interpretable DL techniques have been proposed that can lead to a more trustworthy artificial intelligence (AI) by providing detailed explanations of \textit{why} a model is making a given recommendation. In this vein, Gradient-weighted Class Activation Mapping (Grad-CAM) is a post hoc explainability technique that has been widely applied because it can visually indicate which portions of a model input are most important for a given classification or prediction. However, when using this method, the connection between a given sample's feature importance and the most relevant portions of the training dataset can remain unclear, shielding the user of the AI from the underpinning logic driving it's output. 
\end{abstract}


\begin{highlights}
\item An interpretable DL method allows human users to understand model predictions of bearing health
\item Two-step procedure: health library creation and prediction basis retrieval
\item Representative training samples are selected as prediction basis samples for each test sample
\item Prediction basis samples have similar feature importance distributions to the test sample
\end{highlights}

\end{frontmatter}


\section{Introduction}
\label{sec:Introduction}

Recent years have ushered in a continuous flow of new machine learning research contributions and a steady increase in the number of implementations of machine learning techniques in commercial applications. These advances have been enabled in part by ever-faster computing equipment becoming available that enable cost-effective machine learning deployments. Engineering applications for machine learning techniques are no exception to this trend, as machine learning-enabled health management and fault classification systems are rapidly expanding to monitor and supervise physical assets such as rotating equipment (e.g., bearings and gearboxes) \cite{Wang2019deep, liu2020review}, structural systems (e.g., bridges, buildings, and planes) \cite{Wang2017cost,yiwei2017cost,wang2022structural}, energy infrastructure (e.g., batteries and turbines) \cite{Liu2020machine, Downey2019physics, tsui2015prognostics}, and manufacturing processes \cite{Wang2020machine}. 
%Effectively applying machine learning to complex engineered systems requires a detailed understanding of the operation and optimal health of each system component in order to make conclusions regarding the present or future operating health of that system. 

In this study, we focus on using machine learning to supervise rotating industrial equipment (e.g., pumps, motors, fans, or generators), as these types of assets represent a specific mission-critical class of mechanical system, the failure or which can lead to substantial unplanned downtime, damage to other elements of the equipment, and a significant safety risk \cite{tsui2015prognostics, cerrada2018review}. The criticality of these rotating systems have provided the stimulus for extensive research dedicated to monitoring, classifying, and predicting the health of rotating equipment \cite{Khorasgani2016methodologies, Medjaher2012remaining, Li2015improved}. 

In recent years, deep learning (DL)  - a subset of machine learning that employs layered artificial neural networks to model complex data abstractions - has gained popularity because of its excellent performance across different fields \cite{lou2023recent, shamshirband2021review}. Despite its promising performance, DL models build the input-output knowledge by optimizing model parameters via the gradient backpropagation algorithm. However, the huge amount of parameters inside the DL models makes it too complex to understand the model's decision-making process, which creates challenges for end users of the models interpreting model outputs and determining if results can be trusted in critical applications \cite{Kenny2021explaining, Aas2021explaining}. This challenge is amplified in fields including biomedical engineering and medicine, where scientific ethics demands a high level of trust in machine learning models prior to making medical decisions informed by model outputs \cite{Lee2019explainable, van2022explainable}. Similarly, for safety-critical production tasks or maintenance decisions where rotating equipment failure could lead to millions of dollars in lost revenue, an operator requires far more information from a trustworthy model than a simple prediction result.
%While many engineering applications do not face the same requirement for model interpretability, the need or desire to understand the cause of a models decision remains crucial for end users maintaining equipment, assessing the probability the model prediction is correct, or determining underlying conditions which contribute to the decision.  

Many efforts have been made to improve the performance and subsequent trust that can be placed in the outputs of a DL model. One common approach to improving model trust involves increasing the quality and quantity of training dataset. Unfortunately, generating a sufficient quantity of training data is either cost-prohibitive or physically impossible \cite{Lee2018robust}. Interpretable DL modeling techniques were developed to improve the model's trustworthiness without increasing the amount of training data. By generating an improved model with outputs that are interpretable by the end user, it becomes possible to trust a model that was trained on a smaller but obtainable dataset. %In the wake of this struggle for interpretability, the field of interpretable DL has emerged in an effort to remove model ambiguity and to improve model trustworthiness and interpretability for end users.

%As a data-driven model, DL models learn knowledge from training data by the gradient backpropagation algorithm. However, the huge amount of parameters inside the DL models makes it too complex to understand the model's decision-making process. For safety-critical tasks, end users expect the model to provide prediction results as well as the reason for making such predictions to gain trust in the model \cite{jia2022role}. Grad-CAM has been widely used to provide visual explanations for a trained DL model, and lots of Grad-CAM interpretable DL methods have been developed and implemented \cite{selvaraju2017grad, kim2020bearing, yu2022novel}. Inspired by those methods, the proposed method adopts Grad-CAM to generate a health library containing the required diagnostic knowledge. \textcolor{red}{[This paragraph needs to go to the introduction. - Chao]}

Multiple techniques can be utilized to develop interpretable DL methods, such as guided backpropagation, deconvolution, and class activation maps. Class Activation Mapping (CAM) enhance model explainability by generating activation maps via the weighted sum of the convolutional feature maps before the final output layer \cite{zhou2016learning}. The CAM is applicable for CNNs which do not contain any fully-connected layers. Gradient-weighted Class Activation Mapping (Grad-CAM) was developed as a generalization of CAM that can provide visual explanations to CNNs with fewer restrictions regarding model architecture \cite{kim2020bearing}. This visualization enables the identification of the regions of a sample most important to a trained model's prediction (classification or regression). %and visual explainability techniques can be easily added to CNN models to visualize CAMs, Grad-CAMs, or other variants of visual activation maps overlaid on the predicted sample image. 

Selvaraju et al. implement Grad-CAM as a gradient-based visual explanation which provides a localization map identifying important regions for predicting the sample's class \cite{selvaraju2017grad}. This results in improvements to model training procedures and makes the model output more explainable to end users. Camalan et al. (2021) utilize CAMs to generate heat maps overlaid on test images to identify the location of oral cancer and calculate a patch around the region \cite{Camalan2021convolutional}. %After CAM heatmaps and classifications are determined, these results are fed through an Inceptron-ResNet-V2 model using transfer learning to determine boundaries of the regions most involved in the classification decision making. 
Lee et al. (2018) propose a novel Pyramid Gradient-based Class Activation Mapping (PG-CAM) method that overcomes a lack of highly annotated data by training in a weakly supervised state \cite{Lee2018robust}. %PG-CAM extracts a multi-scale Grad-CAM heatmap which is overlaid onto an image of a potential brain tumor. This approach outperformed traditional Grad-CAM techniques and provides improvements to inconsistent and error-prone manual inspections of patient imaging. 

The aforementioned studies are limited to generating activation heatmaps visualizing important regions of an input sample and still leave challenges for end users looking to fully understand and trust a model's decision-making process. For example, an activation map generated by Grad-CAM can provide end users with an indication of important regions for a particular sample, but this leaves the end user with little context when comparing other samples of the same classification (i.e., how the activation map differs for samples of each classification or how activations may differ between similarly classified samples). 

%Attempting to provide a sample's activation map in addition to maps of similar samples used in model training is a common approach used to provide the context needed to understand what a model deemed important to make a classification as well as to identify if similar classifications exist in the training data. 

A user-friendly interpretable DL method aims to provide training samples that are similar to the test sample during the model evaluation process. Kim et al. (2022) utilize CAM to identify regions of an image most important for classifying chest X-rays \cite{Kim2022accurate}. Then, the model's decision is supported by comparing an activated region of the test X-ray to a library of activation patches generated from training data. From this library a prediction basis is selected that provides images of similarly activated image patches of the same classification. In a similar fashion, Lee et al. (2019) utilize CAM activation maps as an explainable element to a CNN in order to classify hemorrhage conditions within human brain CT scans \cite{Lee2019explainable}. The authors generate a library of training activation maps selected based on a relevance count metric determined for each classification and for activation maps within each layer of the CNN. For a given test sample, prediction basis samples can be selected by calculating the Euclidean distance between the normalized test activation and each library entry. From these distances, the most similar entries are selected to justify the model classification for the end user. 

In this paper, we investigate a novel interpretable DL method applied for the first time in the field of bearing fault diagnosis. First, a CNN model is trained using the training bearing dataset. Then, we calculate each sample's Grad-CAM activation map to generate an activation vector. Each sample's activation vector and its predicted health condition are utilized to create the health library. During the model evaluation process, the proposed method compares the activation vector's similarity of testing samples with each library entry. The entries most similar to the test activation vector are selected as the prediction basis samples. %This collection of training samples utilized as a prediction basis selected using this physics-informed feature is demonstrated to outperform alternative selection procedures, maximizing the usability of the prediction basis and model trustworthiness for an end user. 

\section{Background}
\label{Background}

This section discusses approaches implemented for data pre-processing, DL model design, and Grad-CAM calculation and comparisons. Section \ref{sec:Characteristic_Frequencies} introduces the calculation of physics-informed fault frequencies. Section \ref{sec:Signal_Processing} discusses the signal processing steps that are required to convert the vibration signal from time domain to order domain via envelope analysis and resampling. The CNN architecture is then overviewed in section \ref{sec:CNN_Architecture} with the subsequent calculation of Grad-CAM activation maps discussed in section \ref{sec:Grad-CAM}. 

\subsection{Bearing Fault Characteristic Frequencies}
\label{sec:Characteristic_Frequencies}
Rolling element bearings are key components in rotating mechanical systems. A rolling element bearing contains four main components: an inner race, an outer race, rolling elements (e.g., balls or rollers), and a cage that maintains equal spacing between rolling elements \cite{mcinerny2003basic}. If a defect presents in any bearing component during bearing operation, the contact between the defect area and other components will generate periodic vibrations. Studies have shown that the frequency of that periodic vibration, denoted as bearing fault characteristic frequency, can be calculated based on the geometry of the bearing \cite{Shen2021physics, wang2014rolling}.

This study considers two common bearing fault types: outer race fault and inner race fault. Their related fault characteristic frequencies are calculated and marked as $f_\mathrm{BPFO}$ and $f_\mathrm{BPFI}$, respectively. Bearing characteristic frequencies serve as invaluable features for identifying fault conditions within roller bearings \cite{Rai2016review}. With known parameters, including bearing shaft speed $f_\mathrm{r}$, bearing inner race diameter $d$, outer race diameter $D$, load angle from the radial plane $\phi$, and the number of rollers $n$, $f_\mathrm{BPFO}$ and $f_\mathrm{BPFI}$ can be calculated using the following equations: 

\begin{equation} \label{BPFOfreq}
f_\mathrm{BPFO}=\frac{(n\times f_\mathrm{r})}{2}[1-\frac{d}{D}\times\cos(\phi)]
\end{equation}

\begin{equation} \label{BPFIfreq}
f_\mathrm{BPFI}=\frac{(n\times f_\mathrm{r})}{2}[1+\frac{d}{D}\times\cos(\phi)]
\end{equation}

\subsection{Signal Processing}
\label{sec:Signal_Processing}

Localized bearing faults creating fault characteristic frequencies cause amplitude modulation which results from a combination of the bearing fault frequency as well as the resonance of the entire system \cite{Shen2021physics}.  This reality makes fault frequencies challenging to inspect from raw frequency spectrum data. Data processing is employed to address this challenge and improve usability of frequency data. 

Envelope analysis is a technique capable of demodulating bearing vibration data \cite{Randall2011Rolling,Borghesani2013}. The first step of this process is to transform each time series sample $x(t)$ via a Hilbert transform shown in Eq. (\ref{eq:Hilbert_transform}). This process converts $x(\tau)$, which is the sample at time $\tau$, to generate Hilbert-transformed $\hat{x}(t)$ \cite{Huang1999new}. Envelope analysis is then completed by taking the absolute value of the Hilbert-transformed sample.

\begin{equation} \label{eq:Hilbert_transform}
\hat{x}\left(t\right)=\frac{1}{\pi}\int_{-\infty}^{\infty}{\frac{x\left(\tau\right)}{t-\tau}d\tau}
\end{equation}

To obtain data in the frequency domain, $x(\omega)$, a Fourier transform is completed as shown in Eq. (\ref{eq:Fourier_transform}). This step takes the Hilbert-transformed time domain sample $\hat{x}(t)$ and results in a Fourier-transformed sample in the frequency domain  $f(\omega)$.

\begin{equation} \label{eq:Fourier_transform}
f\left(\omega\right)=\int_{-\infty}^{\infty}{\hat{x}\left(t\right)e^{-j\omega t}dt} 
\end{equation}

As each bearing sample's characteristic frequency is a function of the constant shaft speed, $f_r$, order frequency normalization is utilized to transform the data to be measured in orders of shaft angle. The transformation from frequency to order follows Eq. \ref{eq:Order} where $f$ is a given frequency prior to being transformed to order domain \cite{fyfe1997analysis}.

\begin{equation} \label{eq:Order}
o = \frac{f}{f_r}
\end{equation}

\begin{figure}[htp]
    \centering
    \includegraphics[width=12cm]{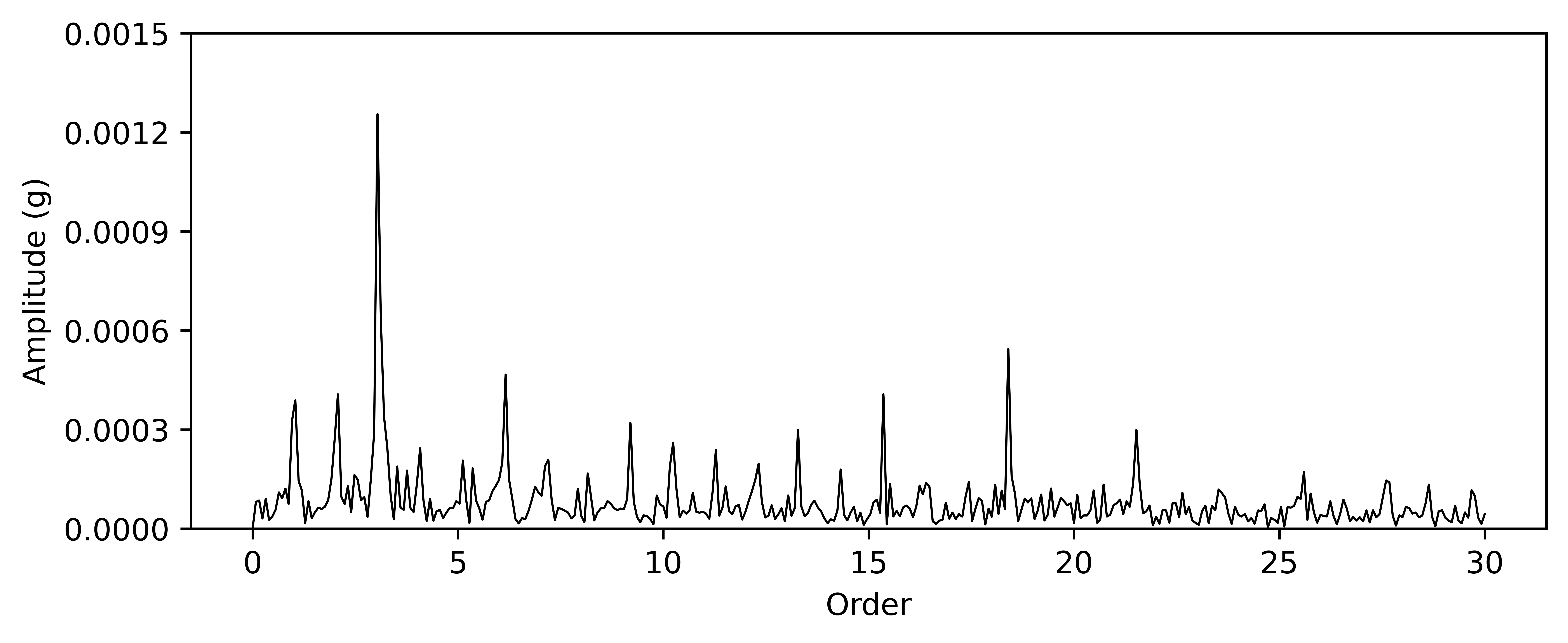}
    \caption{Processed envelope order spectrum.}
    \label{fig:ORSampleNoBands}
\end{figure}

This pre-processed vibration data is now suitable for model training. By working in the frequency domain and completing an envelope analysis, bearing characteristic frequencies can be overlaid on the bearing data. This will also facilitate the identification of class activation values near these characteristic frequencies, which will be discussed later in Sec. \ref{sec:library_creation}. 

\subsection{CNN Architecture for Bearing Health Classification} 
\label{sec:CNN_Architecture}

We utilize a 1D-CNN model to classify bearing health based on pre-processed vibration signals. Each pre-processed vibration signal is in the form of an envelope order spectrum (see an example in Fig. \ref{fig:ORSampleNoBands}). The 1D-CNN model comprises a feature extractor, a global max pooling layer, and a classifier. The feature extractor consists of a series of three convolutional blocks. Each convolutional block is composed of a 1D convolutional layer, a batch normalization layer, and a max pooling layer. The input and output of the $l^\mathrm{th}$ convolutional layer is denoted as $\mathbf{x}^{l}$ and $\mathbf{y}^{l}$, respectively. 

As the first layer inside the $l^\mathrm{th}$ convolutional block, the 1D convolutional layer applies a convolution operation to the input feature map $\mathbf{x}$, generating an output feature map $\mathbf{y}$, expressed as:

\begin{equation} \label{eq:convolution_multiplication}
\mathbf{y}^{l}_{b}=\mathrm{\phi}\left(\sum_{a} \mathbf{x}^{l}_{a}\ast \mathbf{w}^{l}_{ab}\right)
\end{equation}
where $\mathbf{x}^{l}_{a}$ denotes the $a^\mathrm{th}$ channel of the input feature map, $\mathbf{y}^{l}_{b}$ denotes the $b^\mathrm{th}$ channel of the output feature map, $\mathbf{w}^{l}_{ab}$ denotes the 1D convolutional kernel weights (from the $a^\mathrm{th}$ channel of the input feature map to the $b^\mathrm{th}$ channel of the output feature map), $\ast$ denotes a convolution operation, and $\mathrm{\phi}$ is an activation function. Note that the input feature map of the first convolutional layer,  $\mathbf{x}^{1}$, is a vector of vibration amplitude readings from an envelope order spectrum, which can be viewed as a feature vector.

Nonlinear functions are utilized as activation functions in CNNs to give the network the ability to model nonlinear input-output relationships. \cite{hao2020role}. %The activation function selected for use in a model dictates many parameters of model, including preventing vanishing gradient problems \cite{Khagi2022novel}, minimizing computational cost \cite{abdelouahab2017tanh, liew2016bounded}, and enabling gradient based methods by selecting a differentiable activation function \cite{sharma2017activation}. 
Popular nonlinear activation functions, including sigmoid, tanh, and rectified linear unit (ReLU), can be used within CNNs \cite{alma99422496511302432}.  The ReLU activation function shown in Eq. (\ref{eq:ReLU}) is adopted in our work. Note that $z$ in Eq. (\ref{eq:ReLU}) is the scalar input to the ReLU activation function. % as it yielded accurate results and proved more repeatable than alternative activation functions tested.  

\begin{equation} \label{eq:ReLU}
\mathrm{ReLU}(z) =\max{(0,z)}
\end{equation}

%The batch normalization layer is designed to normalize the input across mini-batches. \cite{Ioffe2015batch}. 
In each convolutional block, the batch normalization layer is implemented after the 1D convolutional layer to improve model training stability. Then the output of the batch normalization layer is fed as the input to the 1D max pooling layer. The 1D max pooling layer extracts the maximum feature value within a sub-region by sliding a fixed-length window over its input feature map.

In this study, three convolutional blocks are stacked one by one to extract lower-dimensional features. Following the feature extractor, the global max pooling layer calculates the maximum value within each channel to generate a downsampled feature map \cite{ma2019fine}.
% The final output shape of each CNN block is modified using a max pooling layer. A 1D max pooling layer down samples the output feature maps. Utilizing max pooling ensures each layer produces a feature map that highlights the most present feature, as opposed to the average feature which would be created using an average pooling layer \cite{ma2019fine}. A stride length of 2 is chosen to control the window shifts following:

% \begin{equation} \label{eq:SoftmaxOutputSize}
% Output Size =  \frac{(Input Size-Pool Size+1)}{Strides}
% \end{equation}
As the last component of the CNN, the classifier is designed as a single fully connected layer plus a softmax function. The fully connected layer calculates the weighted sum across all the inputs and adds a bias. Then, the softmax function converts the output of the fully connected layer to class probabilities, i.e., a probability distribution over predicted health classes. Specifically, the softmax function outputs a probability for each health class $c$ with $c = 1,...,C$, where $C$ denotes the number of possible health classes.

\subsection{Grad-CAM Activation Map Visualization}
\label{sec:Grad-CAM}

Grad-CAM visualizations of class activation maps rely on average gradients as weights applied to a map. To calculate a Grad-CAM map for a specific class $c$ out of $C$ possible class labels, the score for class $c$, $y^c$, and the activation map $A_i^k$ for each $i^\mathrm{th}$ element and $k^\mathrm{th}$ filter are retrieved. These values are calculated within a CNN during the back-propagation step as $\frac{\partial y^c}{\partial A_i}$ and are subsequently retrievable without significant additional computational overhead \cite{kim2020bearing}. 

\begin{equation} \label{eq:ImportanceWeights}
\alpha_k^c=\frac{1}{C}\sum_{i}\frac{\partial y^c}{\partial A_{i}^k}
\end{equation}

Utilizing the activation map $A_i$ retrieved from the back-propagation step as well as the importance-weight  $\alpha_k^c$ for class $c$ and filter $k$, the overall class $c$ importance-weight, $L_\mathrm{Grad-CAM}^c$, can be calculated using Eq. (\ref{eq:Grad-CAM}). 

\begin{equation} \label{eq:Grad-CAM}
L_\mathrm{Grad-CAM}^c=\sum_{k}\alpha_k^cA^k
\end{equation}

Activation maps can be generated for each block within a CNN model. Displaying activation maps for different blocks results in varied resolution and detail of the activation map. Later block's feature maps include less noise as a result of passing through additional max pooling layers in each block \cite{li2023multilayer}. As a clearer, readable activation map is desired, the feature map from the last convolutional layer is selected.

\section{Methodology}
\label{sec:Methodology}

\begin{figure}[htp]
    \centering
    \includegraphics[width=\textwidth]{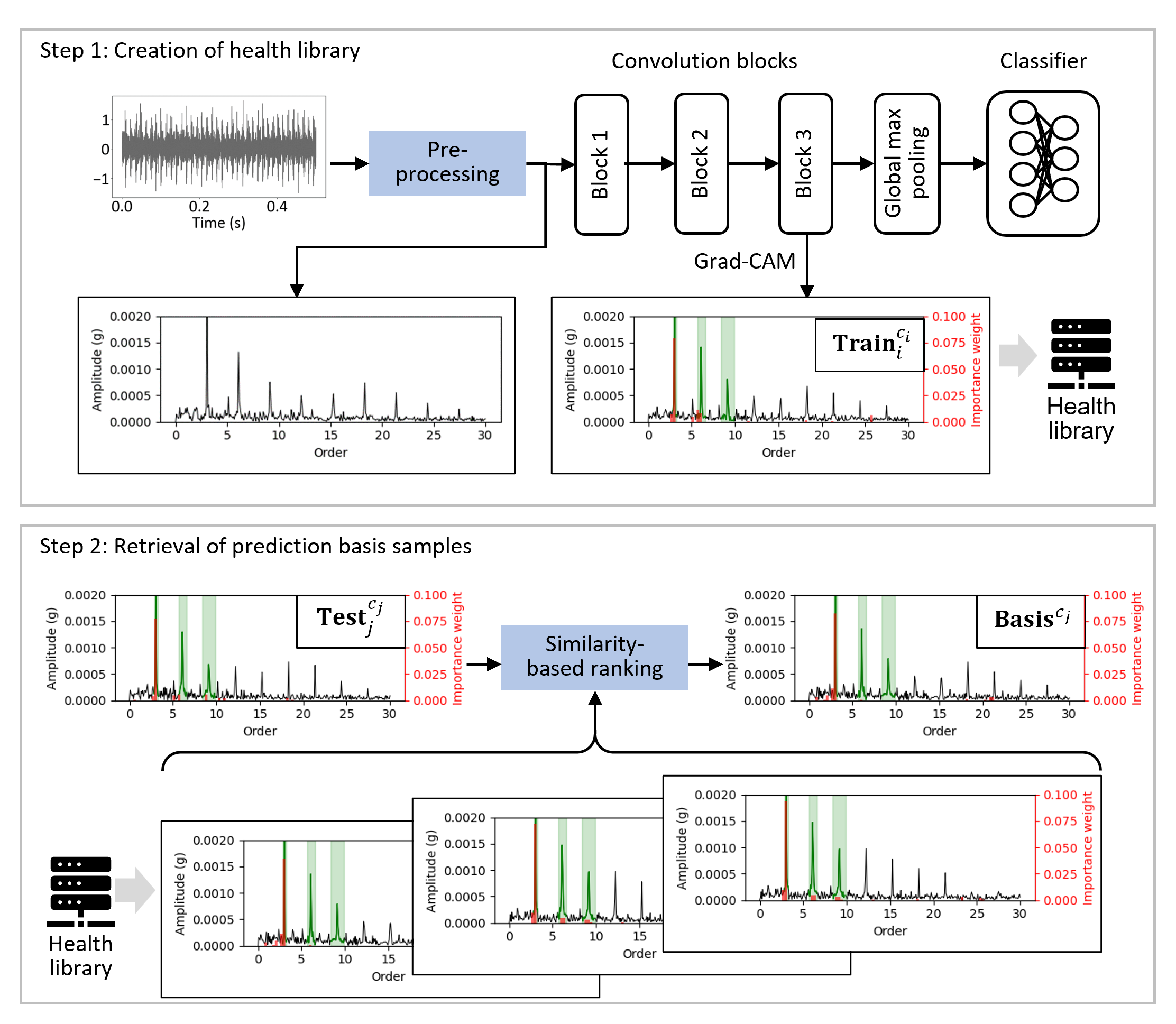}
    \caption{The overview of proposed interpretable DL method}
    \label{fig:Proposed_method}
\end{figure}

The proposed method is a post hoc interpretable DL method applicable to any CNN-based model without modifying the model architecture. An overview of the proposed method is provided in Fig. \ref{fig:Proposed_method}. The proposed interpretable DL method has two major steps: (1) generating a library of candidate prediction basis samples (Sec. \ref{sec:library_creation}) and (2) retrieving prediction basis samples from the library based on the similarity between the feature importance of a test sample and all training samples in the library (Sec. \ref{sec:prediction_basis_retrieval}).

\subsection{Creation of Health Library}
\label{sec:library_creation}

Step 1 of our proposed interpretable DL method constructs a health library based on the class activation maps generated by the last convolutional layer of a trained CNN. The health library is designed as a set with $N_\mathrm{train}$ entries, where $N_\mathrm{train}$ denotes the number of training samples. Each entry consists of an activation vector that reflects the importance of input features and the training sample's health class. Two interpretable DL algorithms are investigated in this study: CAM-Full and CAM-Sub. These two algorithms vary in the activation vector for the health library creation and prediction basis retrieval. 

CAM-Full creates a health library and retrieves prediction basis samples using activation vectors associated with input features (i.e., vibration amplitudes on envelop spectra) across the entire order range $[0, 30]$. CAM-Sub differs from CAM-Full by only considering the feature importance values within the pre-defined frequency sub-bands corresponding to the sample's fault type. The remainder of this subsection describes the design of fault frequency sub-bands.

As described in Sec. \ref{sec:Characteristic_Frequencies}, features located close to the bearing fault characteristic frequencies are important in diagnosing the bearing health. To leverage that domain knowledge, CAM-Sub focuses on the feature importance values associated with input features (vibration amplitudes) near the bearing fault characteristic frequencies. For a given fault type whose health class is ${c}$, the selected input features fall within three frequency sub-bands that are defined using the fault characteristic frequency $f^{c}$ and sub-band width $\epsilon$. These three sub-bands are centered at $f^c$ and it's first two harmonics ($2f^c$ and $3f^c$), respectively. The width of each sub-band is defined as $2\epsilon$ of the sub-band center frequency, and each sub-band extends symmetrically in both directions by $\epsilon$ times the corresponding center frequency. After identifying the order range of the three sub-bands for each fault type, the feature importance (or class activation) values in the sub-bands are chosen to create the CAM-Sub health library.

\begin{figure}[htp]
    \centering
    \includegraphics[width=\textwidth]{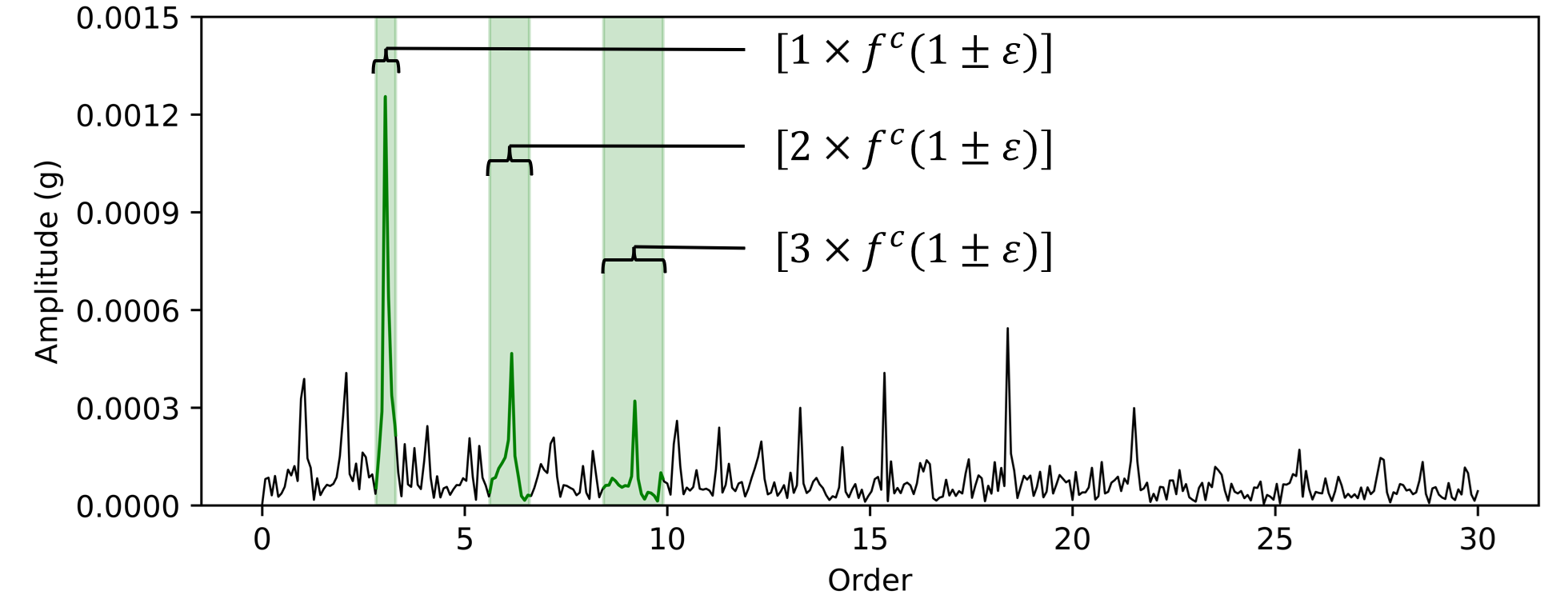}
    \caption{Fault sub-bands (green) overlaid on a pre-processed outer race sample for the bearing outer race frequency $f_\mathrm{BPFO}$ and its first two harmonics.}
    \label{fig:ORSample}
\end{figure}

Figure \ref{fig:ORSample} shows the envelope spectrum of an outer race fault sample, annotated with the three fault frequency sub-bands. The sub-bands include the first outer race fault characteristic frequency and its following two harmonics. Note that the frequency sub-band increases in width (by $2 f^c \epsilon$) for each harmonic. The increasing sub-band width ensures that the first two harmonics of the fault characteristic frequency fall within the respective sub-bands in the presence of errors in the shaft speed measurement. 

\subsection{Retrieval of Prediction Basis Samples}
\label{sec:prediction_basis_retrieval}

Step 2 of our proposed interpretable DL method takes place after a DL model has been trained. Given a test sample, this step calculates the sample's class activation (or feature importance) vector using the CAM-Full or CAM-Sub algorithm described in Sec. \ref{sec:library_creation}, then evaluates the similarity between the class activation vector of this test sample and the class activation vector of each training entry in the health library that has the same health class as the test sample's predicted class. The entries most similar to the test activation vector are selected as the prediction basis samples.

Given a test dataset with $N_\mathrm{test}$ samples, Grad-CAM is used to determine each test sample's class activation vector $\mathbf{TestCA}^{c_j}_j$, where $c_{j}, j = 1,...,N_\mathrm{test}$, denotes the $j^\mathrm{th}$ test sample's predicted class. Then, we select entries in the health library with the same health class $c_{j}$, and their training sample indices form a set of $N_\mathrm{train}^{c_j}$ indices, denoted as $\mathrm{I}^{c_j}$, which satisfies the following condition: $\forall i \in \mathrm{I}^{c_j}, c_i = c_j$. Here, $N_\mathrm{train}^{c_j}$ is the number of training numbers whose health classes are ${c_j}$. The class activation vectors of these $N_\mathrm{train}^{c_j}$ library entries also form a set, denoted as $\mathrm{S}^{c_j} = \{\mathbf{TrainCA}^{c_i}_i\}_{i \in \mathrm{I}^{c_j}}$. Training samples with the health class $c_j$ provide critical information backing the classification of the test sample into $c_j$, assuming that we do not have test samples that fall outside the training data distribution. Considering the health library may contain training data collected from different bearings under different operating conditions, the class activation vector may vary drastically from one training sample to another, even for those with the same health class; thus, it is expected that not all training samples are important to the model prediction on a test sample. To identify prediction basis samples that are highly relevant on a per-test-sample basis, we calculate the similarity between $\mathbf{TestCA}^{c_j}_j$ and $N_\mathrm{train}^{c_j}$ class activation vectors in $\mathrm{S}^{c_j}$. Entries most similar to $\mathbf{TestCA}^{c_j}_j$ are selected as the prediction basis samples. Note again that Sec. \ref{sec:library_creation} describes the derivation of class activation vectors with two algorithms: CAM-Full and CAM-Sub. The key difference between these two algorithms is that CAM-Full selects class activation values associated with vibration amplitudes across the entire order range $[0,30]$, while CAM-Sub concatenates the class activation values within the three pre-defined frequency sub-bands for a given fault type.

Comparisons between two samples (a test sample and an entry in the health library) are achieved by calculating the Euclidean distance between their normalized class activation vectors. The vector normalization is completed using Eqs. (\ref{eq:TestNormalization}) and (\ref{eq:TrainNormalization}), where the class activation vector of the $j^\mathrm{th}$ test sample, $\mathbf{TestCANorm}_j^{c_j}$, and the class activation vector of each library entry, $\mathbf{TrainCANorm}_i^{c_i} \in \mathrm{S}^{c_j}$, are normalized by their respective L2 norms (or Euclidean norms). Utilizing vector normalization minimizes the effect of vector length when measuring the similarity between two vectors. This treatment may be beneficial as our focus is comparing the distributions of relative feature importance rather than the absolute feature importance values.

\begin{equation} \label{eq:TestNormalization}
\mathbf{TestCANorm} _j^{c_j}=\frac{\mathbf{TestCA} _j^{c_j}}{|| \mathbf{TestCA} _j^{c_j} ||}
\end{equation}

\begin{equation} \label{eq:TrainNormalization}
\mathbf{TrainCANorm} _i^{c_i}=\frac{\mathbf{TrainCA} _i^{c_i}}{|| \mathbf{TrainCA} _i^{c_i} ||}
\end{equation}

After the two class activation vectors are L2-normalized, the Euclidean distance between these normalized vectors measures the class activation similarity between the test sample and training entry. This similarity measure takes the following form:

\begin{equation} \label{eq:DistanceCompare}
\mathbf{Dis}_{i,j} = ||\mathbf{TestCANorm}_i^{c_i} - \mathbf{TrainCANorm}_j^{c_j}||
\end{equation}

The Euclidean distance $\mathbf{Dis}_{i,j}$ between a given test sample $j$ and training sample $i$ is calculated for each of the $N_\mathrm{train}^{c_j}$ training entries in the set $\mathrm{S}^{c_j}$. Then, these $N_\mathrm{train}^{c_j}$ selected entries, \{$\mathbf{TrainCA}_{i}^{c_i}\}_{i \in \mathrm{I}^{c_j}}$, are sorted in ascending order of $\mathbf{Dis}_{i,j}$. Finally, the prediction basis samples are retrieved by selecting the top $K$ $(\ll N_\mathrm{train}^{c_j})$ entries with the smallest Euclidean distances to the test sample in the normalized class activation vector space. These representative training samples from the health library with similar distributions of feature importance can be used to justify the prediction of the CNN on the test sample.

\section{Case Study}
\label{sec:case_study}

\subsection{Design of Case Study}

To assess the effectiveness of the proposed interpretable DL method in retrieving trustworthy prediction basis samples, we designed experiments using the open-source dataset from the Paderborn University (PU) Bearing DataCenter. As one of the standard references in the bearing diagnosis field,  the PU bearing dataset has been widely used to evaluate bearing fault diagnosis algorithms. The bearing health condition can be classified as healthy, inner race fault, outer race fault, roller fault, or a combination of faults. The PU dataset generates damaged bearings in two ways: (1) using drilling/electrical discharge machines to introduce artificial damage and (2) performing accelerated life tests to introduce real damage \cite{lessmeier2016condition}. 

\begin{figure}[htp]
    \centering
    \includegraphics[width=\textwidth]{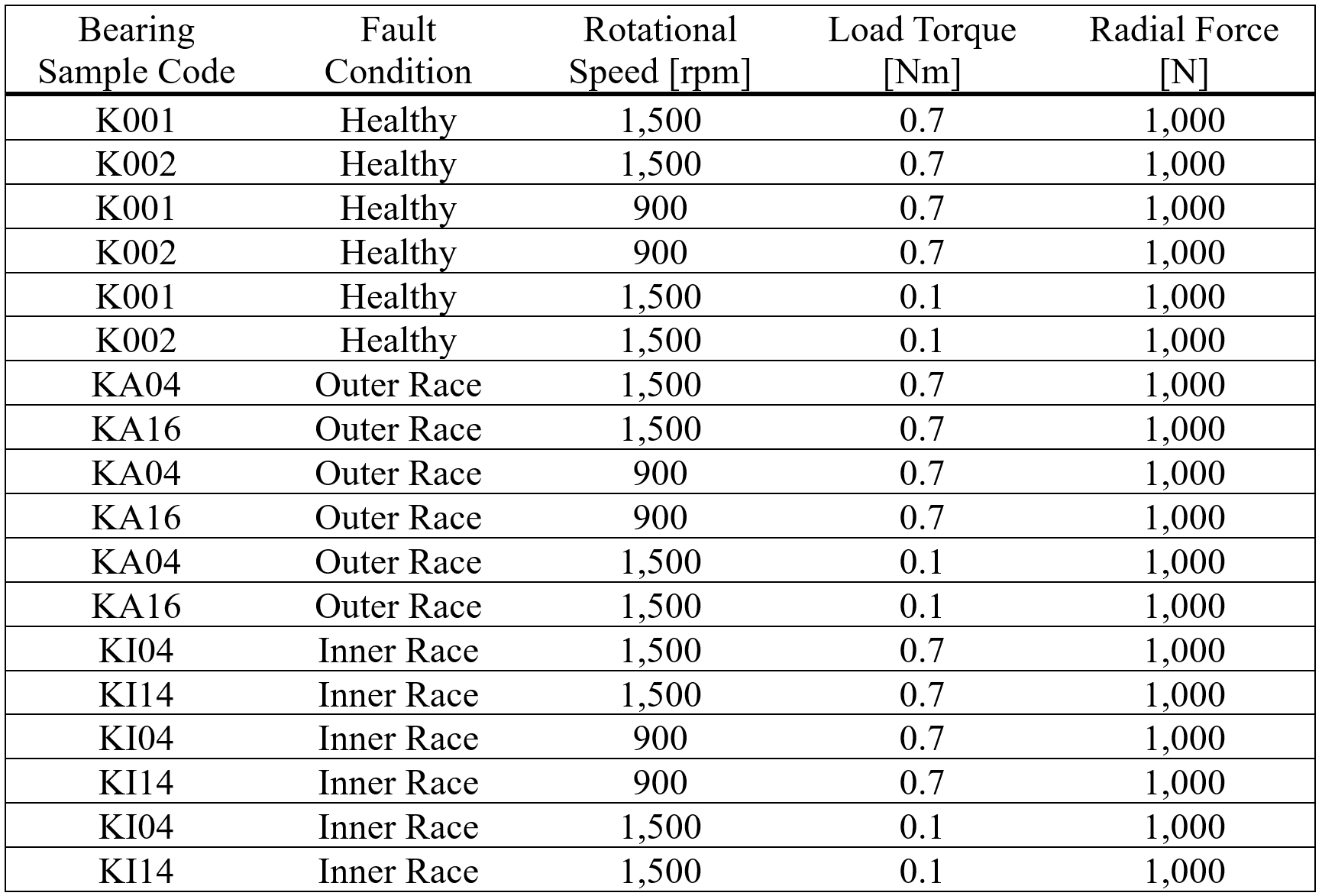}
    \caption{A summary of PU dataset samples selected for this case study.}
    \label{fig:SamplesTable}
\end{figure}

The dataset used in our experiments included bearings subjected to accelerated life tests, shown in Table \ref{fig:SamplesTable}. The data collected from each bearing consisted of three vibration time series samples corresponding to three operating conditions. Each time series sample was cut into smaller samples, with each smaller sample containing 6,400 measurement points. Our signal processing step then converted these vibration amplitude measurements from the time domain into the order domain. We then selected the features within the order range $[0, 30]$ as the model input. After signal processing, 80\% of the samples were used as the training dataset, and the remainder were used as the test dataset.

\subsection{Evaluation of Prediction Basis Selection}

The proposed method aims to provide end users with the model's prediction for each test sample and intuitively show the prediction basis samples, i.e., a set of training samples similar to the test sample that support the model's prediction. The prediction basis samples are expected to reveal information most relevant to the prediction while being easily comprehensible to end users. We conducted experiments to assess the performance of the proposed interpretable DL method in two aspects: (1) the ability to identify prediction basis samples that are important with respect to making accurate predictions on the entire test dataset (Sec. \ref{sec:evaluation_sample_importance}) and (2) the intuitiveness of identified prediction basis samples (Sec. \ref{sec:evaluation_sample_intuitiveness}).

\subsubsection{Importance of Prediction Basis Samples}
\label{sec:evaluation_sample_importance}

We first evaluated the effectiveness of the proposed interpretable DL method in selecting the \emph{most relevant} training samples (prediction basis) that contribute, on average, the most to model predictions on the entire test dataset. The basic idea of our evaluation was to compare the originally trained model with models trained after removing some selected samples from the original training dataset. Specifically, the evaluation was carried out in the following three steps:
\begin{enumerate}
    \item Given each test sample, we used Grad-CAM to compute its feature importance. We then calculated the Euclidean distance between the feature importance vector of the selected test sample and the saved importance vector of each training sample in the health library. We repeated the above process for each test sample to obtain the Euclidean distance between each pair of test and training samples, generating a distance matrix $\mathbf{Dis}$ of size $N_\mathrm{train}\times N_\mathrm{test}$, where $\mathbf{Dis}_{i,j}$ denotes the Euclidean distance between the $i^\mathrm{th}$ training sample and $j^\mathrm{th}$ test sample, and $N_\mathrm{train}$ and $N_\mathrm{test}$, respectively, denote the numbers of training and test samples. 
    \item For each training sample, we computed the average of the distances of the training samples to the test samples with the same label $\mathbf{AvgDis}_{j}=\frac{1}{N_\mathrm{train}}\sum_{i=1}^{N_\mathrm{train}} \mathbf{Dis}_{i,j}$. This average distance measures the importance of a training sample to model predictions on test data, on average, across all test samples combined. Then, we ranked the training samples according to their average distances to the test data (i.e., their overall importance to the predictions on the test samples), removed the top $K\%$  ($K = 10, 15, 20, ..., 45$) training samples, and trained a new DL model with this newly formed, reduced training dataset.
    \item The third and last step evaluated the newly trained model using the test dataset (original, unchanged) and calculated the accuracy difference between the newly trained and original models. 

\end{enumerate}

Our intuition was that removing training samples of higher importance is expected to cause a more significant decrease in model accuracy. If the Euclidean distance metric in Eq. (\ref{eq:DistanceCompare}) allows for effectively identifying high-importance training samples, we expect a more rapid accuracy decrease than randomly removing training samples. Note that the proposed interpretable DL method selects prediction basis samples according to this Euclidean distance metric. If this metric proves effective, the selected prediction basis samples will likely contain more relevant information that contributes to correctly classifying the test sample's health. The baseline in our comparison was a random selection method, denoted as ``Random", which randomly selects a certain fraction of training samples as the prediction basis for sample removal and model re-training. To account for the run-to-run variation in the model training process, we independently re-trained ten new models for each of the three approaches (i.e., Random, CAM-Full, and CAM-Sub). 

\begin{figure}[!h]
    \centering
    \includegraphics[width=\textwidth]{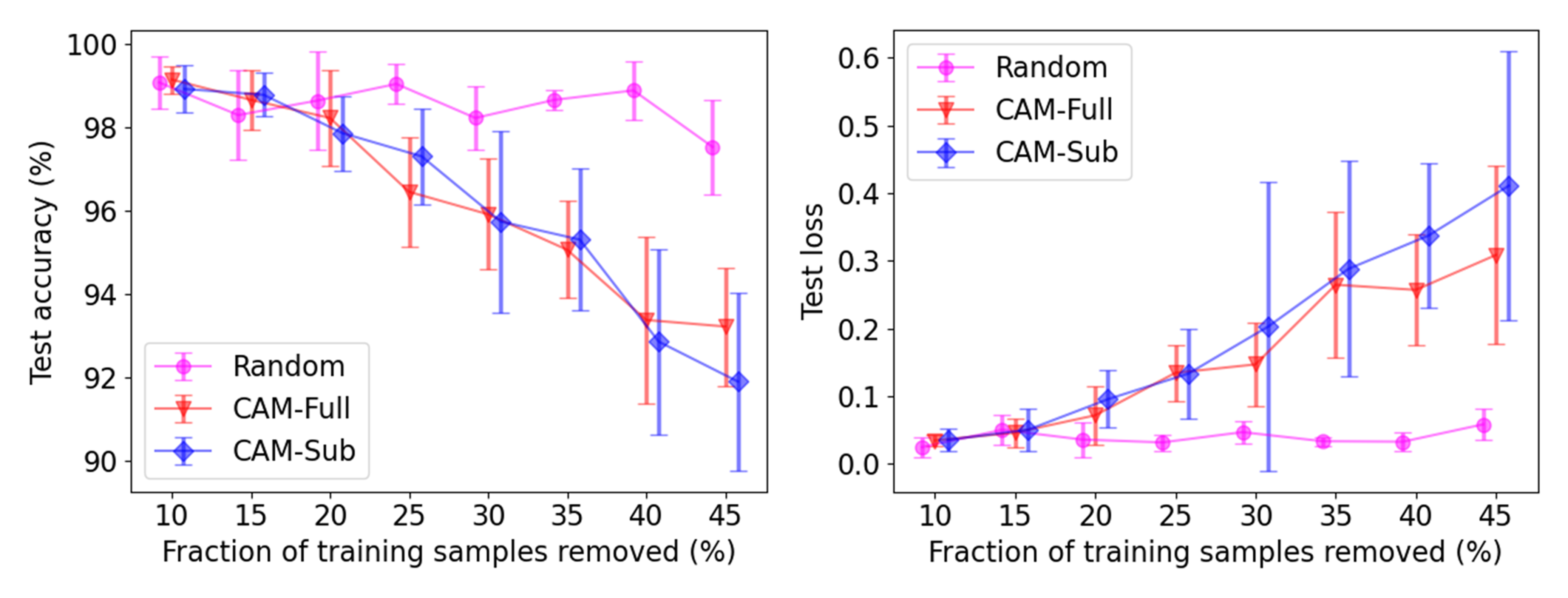}
    \caption{Test accuracy and loss of re-trained models vs. fraction of training samples removed. A more rapidly decreasing trend of accuracy or increasing trend of loss indicates a more effective ranking of training samples according to their overall importance. The error bars for Random and CAM-Sub at each fraction number of interest are offset slightly and horizontally for ease of visualization.}
    \label{fig:change_of_acc}
\end{figure}

The comparison results are summarized in Fig. \ref{fig:change_of_acc}. These two plots visualize how the classification accuracy (Fig. \ref{fig:change_of_acc}(a)) and loss (Fig. \ref{fig:change_of_acc}(b)) on the test dataset change as the fraction of removed training samples increases. Suppose the Euclidean distance metric in Eq. (\ref{eq:DistanceCompare}) is a good measure of training sample importance. In that case, the classification accuracy (loss) should decrease (increase) monotonically and rapidly (at least initially) with the fraction of removed high-importance samples. This seems to be the case for CAM-Full and CAM-Sub, as shown in Fig. \ref{fig:change_of_acc}. The decreasing (increasing) trend of the classification accuracy (loss) by CAM-Sub seems slightly more rapid than that by CAM-Full, suggesting that CAM-Sub performed marginally better than CAM-Full in identifying important training samples. This better performance by CAM-Sub is also evidenced by another observation that model ``re-training" after removing important training samples identified by CAM-Sub exhibited larger run-to-run variation than CAM-Full and Random. In other words, removing prediction basis samples caused increases in the instability of model training with respect to the test dataset, indicating these removed training samples are highly important to ensure consistent accuracy on the test samples in the presence of randomness in the model training process. In contrast, the random selection method that does not consider sample importance produced an almost flat curve for accuracy and loss. This observation provides evidence that looking at the similarity between a test sample’s class activation and a training sample’s class activation offers a meaningful way to measure the importance of the training sample to the prediction for the test sample. 

So far, we have analyzed how the overall classification accuracy and loss change as functions of the fraction of training samples removed. Additionally, we examined a detailed summary of classification results in the form of a confusion matrix. Given a confusion matrix $\mathbf{C}$, the entry $\mathbf{C}_{i,j}$ is equal to the number of samples with true label $i$ classified as label $j$. 
\begin{figure}[!h]
    \centering
    \includegraphics[width=\textwidth]{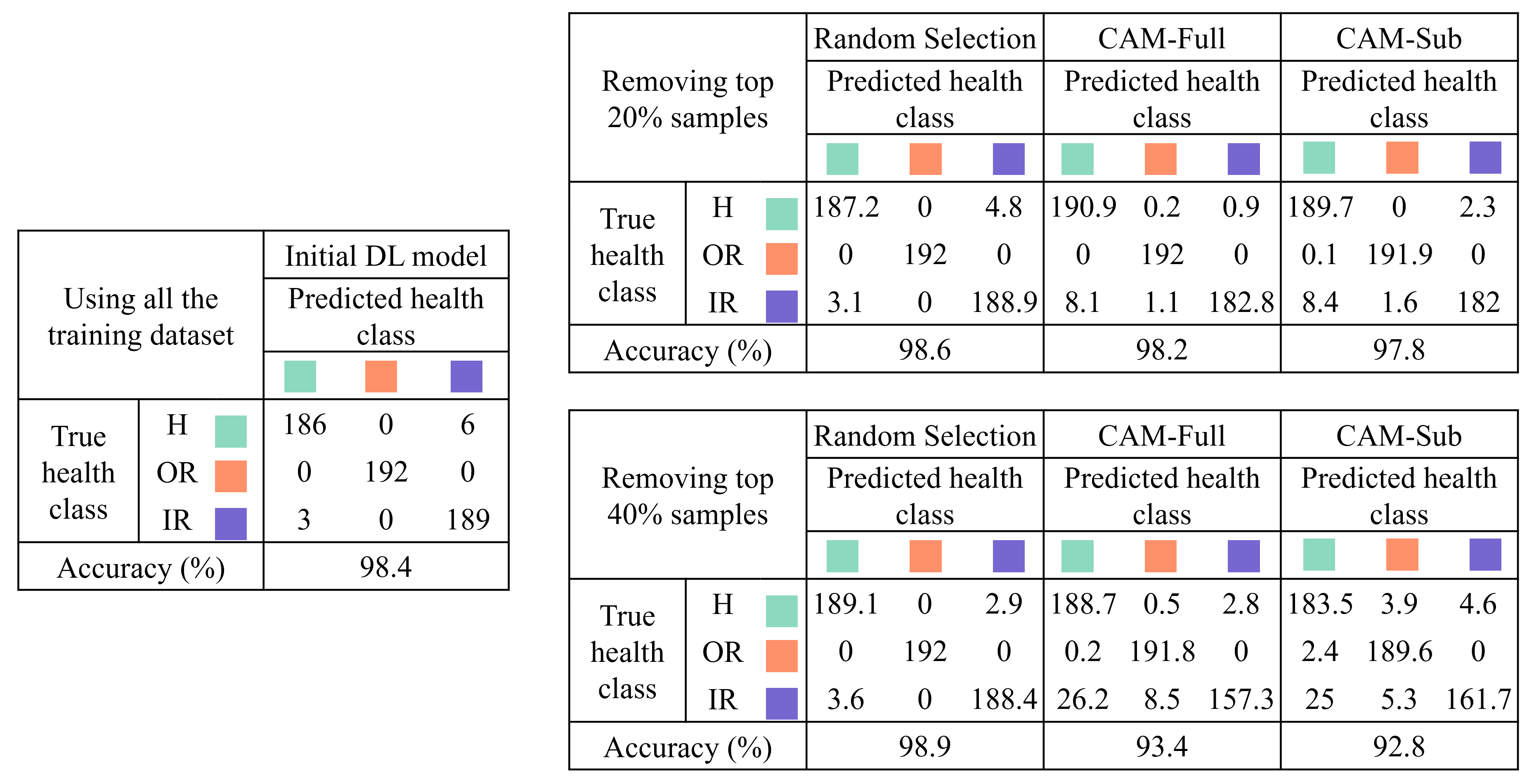}
    \caption{Confusion matrices produced by DL models trained using varying quantities of data. Three methods (Random, CAM-Full, and CAM-Sub) were implemented to remove 20\% and 40\% of all training samples to produce the confusion matrices on the right-hand side.}
    \label{fig:Change_of_CM}
\end{figure}
The confusion matrices produced by the three methods when removing 20\% and 40\% of all training samples are summarized in Fig. \ref{fig:Change_of_CM}. No significant changes in the models' performance were observed after removing the top 20\% of samples from the training dataset.

After removing the top 40\% of samples from the prediction basis pool, the models developed by CAM-Sub and CAM-Full yielded substantially lower prediction accuracy on inner race fault samples. 
Compared with the CAM-Full models, the CAM-Sub models exhibited more noticeable decreases in the classification accuracy on the healthy and outer race fault samples. Considering the training samples removed consisted of healthy, inner race fault, and outer race fault samples, this performance degradation indicates CAM-Sub outperformed CAM-Full in identifying training samples of high importance with respect to the other two health classes (i.e., healthy and outer race fault).

\subsubsection{Intuitiveness of Prediction Basis Samples}
\label{sec:evaluation_sample_intuitiveness}

The second aspect of our method evaluation focused on how intuitive the prediction basis samples were and how well these samples could help end users interpret the model's decision-making process. We selected a representative outer race fault test sample and an inner race fault test sample, and used CAM-Full and CAM-Sub to generate the top four prediction basis samples for each selected test sample.

\begin{figure}[htp]
    \centering
    \includegraphics[width=\textwidth]{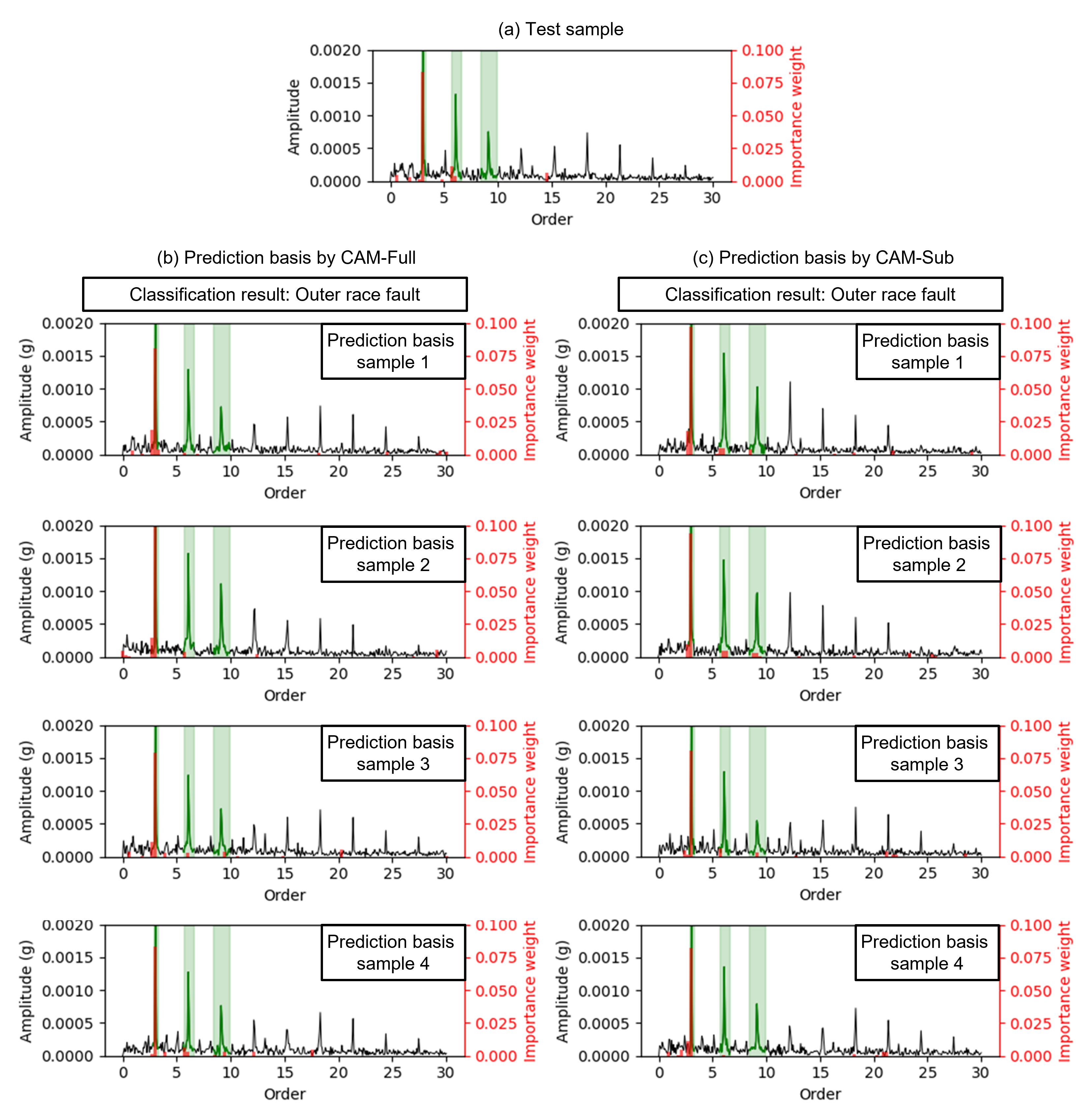}
    \caption{Envelop spectra of an outer race fault test sample (a) and its four prediction basis samples retrieved by CAM-Full (b) and CAM-Sub (c)}
    \label{fig:Results_OF}
\end{figure}

Let us first look at the outer race fault test sample. Figure \ref{fig:Results_OF} shows the envelope order spectrum of the test sample (top row) and the prediction basis samples CAM-Full and CAM-Sub retrieved from the health library (four bottom rows), all overlaid with the feature importance by Grad-CAM. High amplitude peaks can be observed at orders 3.05, 6.10, and 9.15, which are the outer race fault frequency and its first two harmonics. The DL model correctly classified the bearing health condition for this selected test sample. Both CAM-Full and CAM-Sub retrieved prediction basis samples with similar activation maps to the test sample.
These prediction basis samples would allow human users to understand the model's decision-making process, particularly its justification for this prediction. Such understanding can be obtained by comparing the fault-signature similarities between the test and prediction basis samples in the order ranges of interest (e.g., the frequency sub-bands for the outer race fault).

\begin{figure}[htp]
    \centering
    \includegraphics[width=\textwidth]{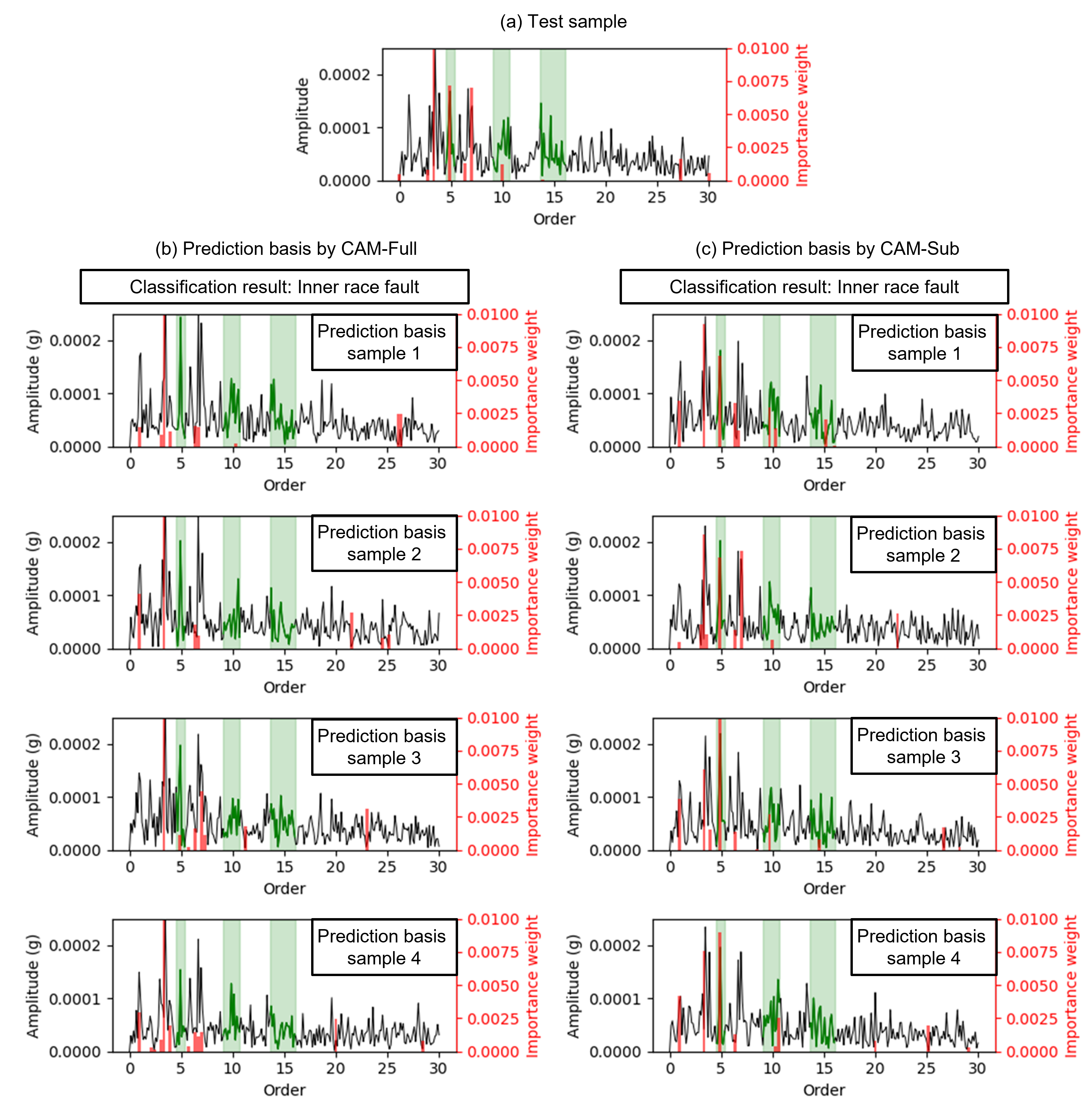}
    \caption{Envelop spectra of an inner race fault test sample (a) and the prediction basis samples retrieved by CAM-Full (b) and CAM-Sub (c)}
    \label{fig:Results_IF}
\end{figure}

We can make a similar conclusion on the intuitiveness of prediction basis samples when looking at the inner race fault test sample, as shown in Figure \ref{fig:Results_IF}. The main difference is that compared to the outer race fault sample shown in Fig. \ref{fig:Results_OF}, the inner race fault sample contained more complex fault signatures, for example, peaks outside of the inner race fault-related sub-bands and weaker peaks. This observation is consistent with prior literature and may be attributed to the fact that the inner race is not stationary like the outer race \cite{cong2013vibration}. CAM-Full selected prediction basis samples according to the Euclidean norm of the vector difference of the entire activation map. Although the four selected training samples were supposed to be the closest to the test sample in terms of the normalized activation (or feature importance) vector, the importance weight distributions of these training samples appeared different from that of the test sample. In particular, the DL model assigned high weights to vibration amplitudes at around order 4.9 (the first fault frequency sub-band), i.e., the inner race fault frequency, for the test sample (see Fig. \ref{fig:Results_IF}(a)), while in contrast, very low importance weights were associated with features in the first fault frequency sub-band for the prediction basis samples retrieved by CAM-Full. Looking at the Grad-CAM outputs of the top four prediction basis samples by CAM-Sub, we observe strong activation (high feature importance) at around order 4.9. This observation suggests that CAM-Sub selected training samples with similar class activation maps to the test sample's class activation map in the order ranges of high relevance to the inner race fault. The results are expected as CAM-Sub selects prediction basis samples according to the similarity of (normalized) importance weight vectors within fault frequency sub-bands. Overall, the prediction basis samples retrieved by CAM-Sub make more intuitive and physical sense than those by CAM-Full, at least for this test sample.

%A panel visualize the Algorithm's output
%Model's prediction results, the L2- distance, selected prediction basis samples.

\section{Conclusion}
\label{section6}

In this study, we have proposed an interpretable DL technique capable of creating a prediction basis for each test sample by ranking the 1D activation maps of training samples. These prediction basis samples are a subset of the training dataset deemed most relevant to the DL model's prediction. Our work furthers our understanding of how to leverage Grad-CAM activation maps to enhance the interpretability of convolutional neural networks (CNNs) in bearing fault classification tasks. A unique benefit offered by our method is an extension of model interpretability from evaluating feature importance (Grad-CAM) to evaluating training sample importance. This extension was made possible by comparing the (normalized) class activation vector of a test sample with those of training samples across the entire frequency range (CAM-Full) or pre-defined fault frequency sub-bands (CAM-Sub). This comparison results in a prediction basis set of training samples (or simply prediction basis samples)  that are expected to contribute the most to the mode's prediction on the test sample. The effectiveness of both CAM-Full and CAM-Sub was evaluated using the Paderborn University (PU) Bearing DataCenter's experimental dataset and validated against randomized sample selection. Our future work will examine ways to account for model uncertainty in retrieving prediction basis samples and investigate model interpretability when applying deep learning to predict bearing remaining useful life. 

\section*{Acknowledgments}
This work was supported in part by the U.S. National Science Foundation under Grant IIP-2222630. Any opinions, findings, or conclusions in this paper are those of the authors and do not necessarily reflect the sponsor’s views. We also thank Dr. Venkat Nemani and Dr. Mohammad Behtash for the helpful discussions on interpretable DL for bearing health monitoring. 

\bibliographystyle{elsarticle-num} 
 \bibliography{XAIBib.bib}





% \end{thebibliography}
\end{document}